\documentclass[letterpaper]{article} 
\usepackage[]{aaai24}  
\usepackage{times}  
\usepackage{helvet}  
\usepackage{courier}  
\usepackage[hyphens]{url}  
\usepackage{graphicx} 
\usepackage{natbib}  
\usepackage{caption} 
\usepackage{amsmath}

\usepackage{algorithm}
\usepackage{algorithmic}

\usepackage{newfloat}
\usepackage{listings}
\DeclareCaptionStyle{ruled}{labelfont=normalfont,labelsep=colon,strut=off} 

\lstdefinestyle{mystyle}{
    basicstyle=\fontsize{8}{10}\selectfont\ttfamily, 
    breaklines=True, 
    frame=single, 
    captionpos=t 
}

\floatstyle{ruled}
\newfloat{listing}{tb}{lst}{}
\floatname{listing}{Listing}
\title{Neural Architecture Codesign for Fast Bragg Peak Analysis}
\author{
    Luke McDermott \textsuperscript{\rm 1},
    Jason Weitz \textsuperscript{\rm 1},
    Dmitri Demler \textsuperscript{\rm 1},\\
    Daniel Cummings\textsuperscript{\rm 2}, 
    Nhan Tran\textsuperscript{\rm 3,4},
    Javier Duarte\textsuperscript{\rm 1}
}
\affiliations{
    \textsuperscript{\rm 1} University of California, San Diego,\textsuperscript{\rm 2} Intel Labs,\\
    \textsuperscript{\rm 3} Fermi National Accelerator Laboratory,
    \textsuperscript{\rm 4} Northwestern University\\
    lmcdermo@ucsd.edu, jdweitz@ucsd.edu, ddemler@ucsd.edu,\\
    daniel.j.cummings@intel.com, ntan@fnal.gov, jduarte@ucsd.edu
}

\pubnote{FERMILAB-CONF-23-0813-CSAID-PPD}

\begin{document}

\maketitle

\begin{abstract}
We develop an automated pipeline to streamline neural architecture codesign for fast, real-time Bragg peak analysis in high-energy diffraction microscopy.
Traditional approaches, notably pseudo-Voigt fitting, demand significant computational resources, prompting interest in deep learning models for more efficient solutions.
Our method employs neural architecture search and AutoML to enhance these models, including hardware costs, leading to the discovery of more hardware-efficient neural architectures.
Our results match the performance, while achieving a 13$\times$ reduction in bit operations compared to the previous state-of-the-art.
We show further speedup through model compression techniques such as quantization-aware-training and neural network pruning.
Additionally, our hierarchical search space provides greater flexibility in optimization, which can easily extend to other tasks and domains.
\end{abstract}

\section{Introduction}
With the rapid advance of machine learning tools, material science researchers have significantly enhanced their experimental methodologies and analysis.
Particularly, locating diffraction peak positions for X-ray diffraction-based microscopy proposes an interesting challenge.
Common methods require large inference costs, thus sparking work in building lightweight deep learning models as approximators.
\citet{BraggNN} laid the foundation for this work, training a transformer-like CNN on pseudo-Voigt fitting predictions~\citep{pseudo_Voigt}, using a diffraction scan from an undeformed bi-crystal gold sample~\citep{dataset}.
This work has extensively been used, combined with \citet{edgeBragg}, to integrate BraggNN with remote data centers for faster model retraining and deployment, enhancing real-time data analysis.
Furthermore, \citet{OpenHLS} heavily optimized how the model is compiled and deployed in hardware, providing a significant speed up in their framework.
OpenHLS compiles BraggNN into a low-latency, register-transfer level format, which increases model throughput.
While these methods lay the foundation for hardware optimization in high-energy diffraction microscopy~\citep{HEDM}, there is considerable room for improvement in the design of the neural architecture. 

Our pipeline incorporates neural architecture search (NAS)~\citep{NAS_survey}, hyperparameter optimization (HPO), and model compression techniques, including pruning~\citep{pruning_survey} and quantization-aware training~\citep{quantization_survey}.
Deep learning models are optimized over a hierarchical search space, selecting high-performing lightweight models with evolutionary strategies.
This automated pipeline streamlines neural architecture codesign (NAC) from model to hardware optimization.
We demonstrate state-of-the-art performance for psuedo-Voigt fitting estimation over previous work, while requiring 13$\times$ less bit operations (BOPs)~\cite{bops} than \citet{BraggNN} for inference.

Our methodology employs a modular approach, based on open-source packages, allowing for flexible and dynamic construction.
This adaptability facilitates the integration or alteration of network components, allowing for bespoke search spaces on new tasks.
We believe in democratizing edge AI for ML practitioners in science domains, especially on specialized hardware such as field-programmable gate arrays (FPGAs).
To carry out this mission, our open-source pipeline will be released in the camera-ready version.

\section{Related Work}
\begin{figure*}[t]
\centering
\includegraphics[width=\textwidth]{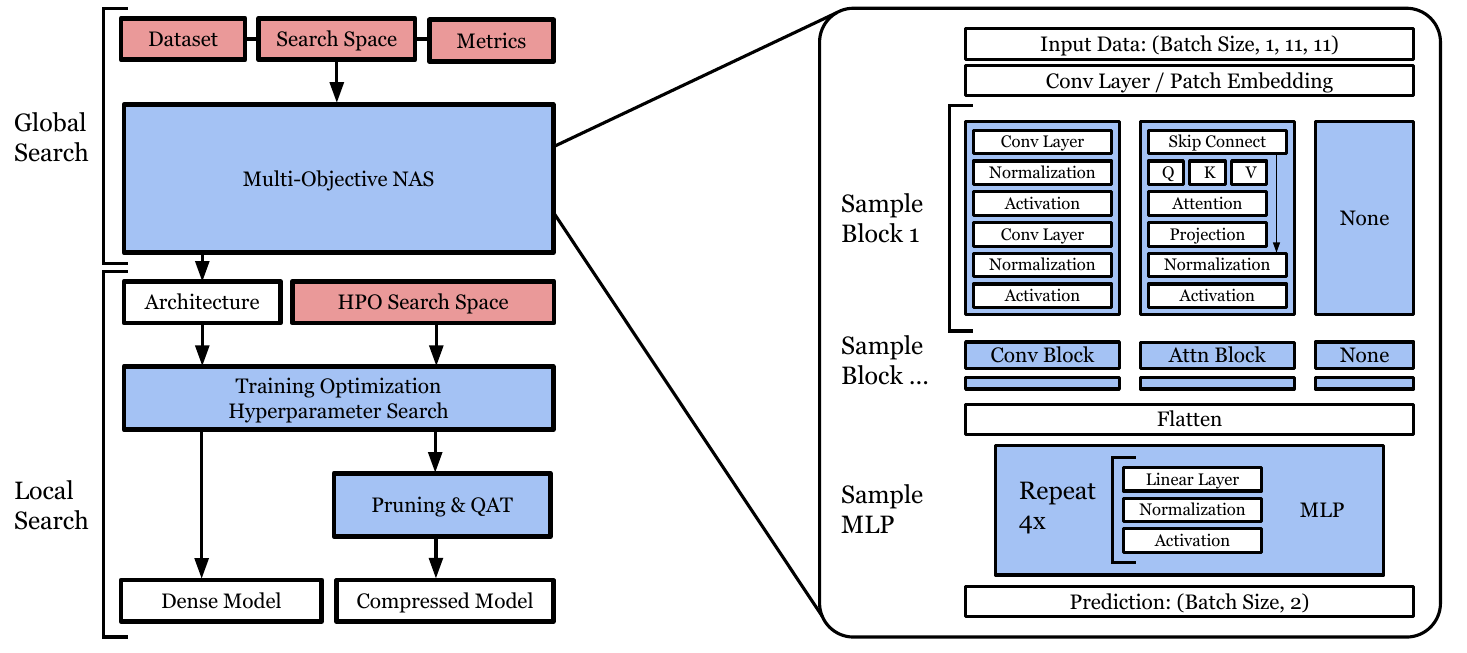} 
\caption{Our automated pipeline for neural architecture search.
Red components are human inputs, white as outputs, and blue as search processes.
The right side demonstrates the template of each candidate architecture in our search space.
Each subcomponent of the blocks also contain hyperparameters to optimize.}
\label{Pipeline}
\end{figure*}
\paragraph{Neural Architecture Search Background.} 
NAS aims to optimize the structure of neural networks for specific tasks and objectives.
This includes searching over network sizes or even constructing completely different model classes.
There are three critical components: \textit{search space, search strategy, and architecture evaluation}.
The search space determines the potential architectures that can be sampled~\citep{search_space1}.
While a narrow search space can be heavily biased, a large one is extremely difficult to properly explore, necessitating a delicate balance.
This space is explored by sampling architectures and evaluating them across our objectives.
Instead of evaluating high-cost objectives, such as network performance, researchers often use a proxy for such objectives like partial training or even zero-cost methods~\citep{zero_cost_nas}.
After validating, the search strategy will update its beliefs and sample again.
Various strategies for this exist such as Bayesian optimization (BO), evolutionary algorithms, or reinforcement learning.
Each has their own use case; for instance, BO methods tend to struggle with high combinatorial categorical hyperparameters, prompting the use of genetic algorithms instead \cite{cai2020once}.
However, BO, specifically tree-structured Parzen estimators (TPEs)~\citep{TPE} perform exceptionally well on continuous hyperparameter optimization tasks where sample efficiency is important~\citep{Optuna}.  
In this paper, we utilize NSGA-II~\citep{NSGA}, a genetic algorithm, for global search due to our discrete and categorical search space, but use a TPE for the continuous hyperparameters in training optimization.

\paragraph{Model Compression Background.}
Other than finding an efficient model class, the small structures of a model can be optimized through neural network pruning and quantization~\citep{modelcompression}.
Pruning aims to remove superfluous parameters in our model, and these algorithms can be categorized by the type of structure they remove~\citep{pruning_survey}.

Structured pruning removes weights associated to larger structures in the network, such as neurons, channels, or attention heads, while unstructured pruning removes individual weights with no specific structure requirements.
Unstructured pruning leads to sparse matrices, which can often limit gains in actual inference speed on GPUs despite the reduced number of parameters.
However, on more versatile or flexible hardware like FPGAs and CPUs, unstructured pruning can provide significant speed up with a negligible drop in performance.
While structured pruning provides definitive speed ups on general hardware, it can lead to a larger decline in performance.
To get around this, newer hardware supports $N:M$ or mixed sparsity~\cite{nvidiaNM}, such as Nvidia's A100 that supports 2:4 sparsity, which alleviates the need to prune entire rows or columns as done with dropping neurons or filters.
Therefore, the choice of pruning algorithm is closely tied to the target hardware for deployment.

Researchers also employ quantization to reduce the number of bits needed to represent the weights or activations.
Like pruning, quantization can be done post-training or during training with quantization-aware training (QAT)~\cite{QAT}.
With QAT, the weights are quantized on the forward pass, but use full-precision gradients on the backward pass, allowing for further fine-tuning of the low-bit representations.
The effectiveness of quantization is heavily dependent on hardware support of low-bit data types.
This framework aims to eventually support deployment on FPGAs, which can support sparse operations and a wide range of reduced precision data types, thus unstructured pruning with sub-8 bit QAT is used.

\section{Method}
We combine neural architecture search, hyperparameter optimization, quantization-aware-training, and neural network pruning all into one automated pipeline, organizing these methods into two stages: \textit{global and local search}.
This pipeline is demonstrated in Fig.~\ref{Pipeline}.
The hierarchical search space is constructed with similar principles as the baseline architectures.
Both \citet{BraggNN} and \citet{OpenHLS} use transformer-like CNNs, essentially replacing the traditional linear layers transformers with convolutional layers and adding an MLP classifier.

\subsection{Global Search}
\paragraph{Search space.}
In this space, each candidate architecture is defined by a feature extractor of up to 3 sequential blocks and a final MLP classifier block. 
At the top level of the hierarchy, block types for each of these 3 layers are sampled from either a conv block, a conv-attention block, or a placeholder block, allowing us to sample shorter networks if needed.
Both baselines can be constructed in this search space with a single conv-attention followed by a traditional conv block.
Within these modules search is performed across block-specific parameters such as normalization methods, activation functions, channel dimensions, etc.
Despite this space being extremely large---on the order of $10^{23}$---possible architectures, the search process is eased with the hierarchical structure and conditional sampling.
In the future, we plan to use methods similar to Monte Carlo tree search to maximally exploit this hierarchy.
This search space is illustrated in Fig.~\ref{Pipeline} on the right and provide an exhaustive description of this space in the Appendix.

\begin{figure}[t]
\centering
\includegraphics[width=\columnwidth]{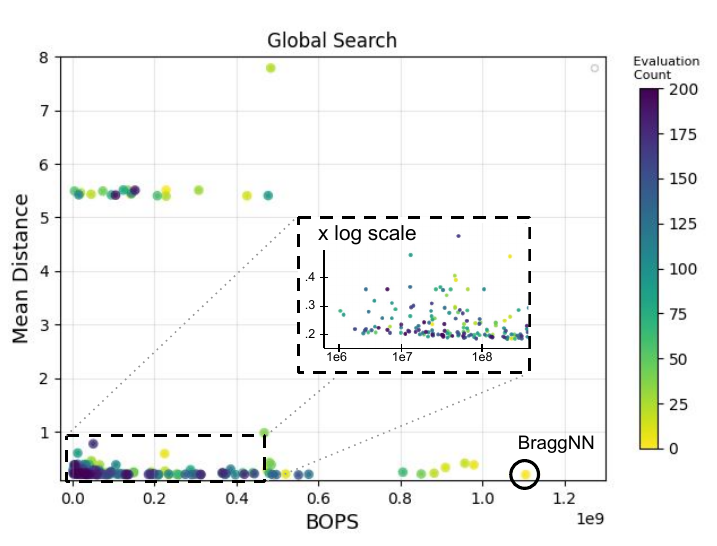} 
\caption{Performance of global search on the validation set over time.
Darker markers represent later trials.
We evaluate each architecture on mean distance and bit operations (BOPs), demonstrating convergence to a pareto front.
Previous work is plotted under the same partial training procedure as the rest of the samples.}
\label{GlobalSearch}
\end{figure}

\paragraph{Search and Evaluation.}
To perform this search, NSGA-II~\citep{NSGA} with population size of 20 is used to optimize our discrete multi-objective problem over 200 trials. NSGA-II minimizes the number of bit operations used at inference and minimizes the distance between our models prediction and the center given by pseudo-voigt fitting on the validation set.
BOPs is used as it generalizes well across hardware and data types rather than FLOPs (floating point operations).
Our models are evaluated with partial-training, using scores after 50 epochs rather than 300 epochs as with full-training.
Our performance on the validation set and model efficiency over each trial in the search process is shown in Fig.~\ref{GlobalSearch}.

\subsection{Local Search}
\begin{figure}[t]
\centering
\includegraphics[width=\columnwidth]{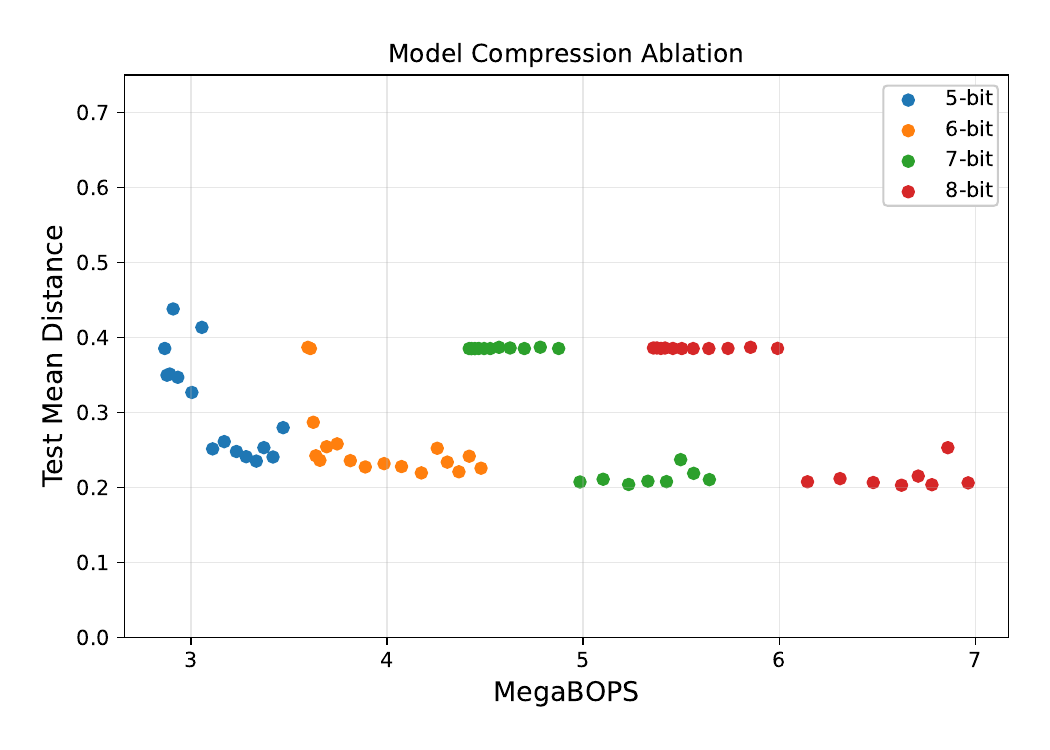}
\caption{Performance versus efficiency for different QAT precisions across pruning.
We record a point at each pruning iteration for a QAT run.}
\label{modelcompression}
\end{figure}

\paragraph{Training Optimization.}
After finding a high performing model class with global search, we now perform a hyperparameter optimization of the training procedure via tree-structured Parzen estimation.
These hyperparameters consist of learning rate, learning rate schedule, weight decay, etc.
This model is reported before any model compression as the NAC base architecture.
For a fair comparison, training the BraggNN architecture is also optimized with 100 trials each.
\paragraph{Model compression.}
To improve the efficiency even further, quantization-aware training can be paired with neural network pruning.
Figure~\ref{modelcompression} illustrates the performance trade off as we quantize to different precisions and sparsify our model.
For each bit precision, iterative magnitude-based pruning is performed with quantization-aware training using Brevitas~\citep{brevitas} inside the inner loop, removing 20\% of the parameters each iteration.
The 7-bit quantized model can match the performance of the dense model at up to 80\% sparsity, resulting in 17$\times$ less BOPs than the uncompressed base model.
In the future, we plan to deploy this on FPGAs with frameworks like hls4ml~\citep{hls4ml} to achieve even better speed up, similar to the work in \citet{OpenHLS}.

\begin{figure}[t]
\centering
\includegraphics[width=\columnwidth]{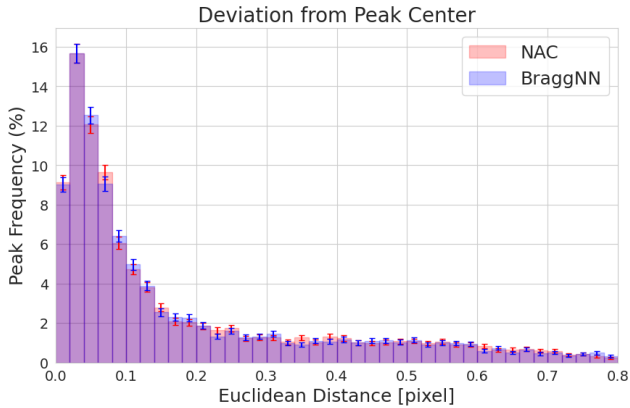}
\caption{Distribution of Euclidean distance difference between peak centers using NAC and BraggNN relative to the pseudo-Voigt fit.}
\label{euclideandist}
\end{figure}

\section{Results}
\begin{table}[t]
\small
\centering
\begin{tabular}{l | c | c | c } 
\hline
Model & Mean distance & MBOPs & Parameters\\
\hline
\hline
BraggNN & 0.202 & 1,106 & 45,274 \\
NAC (ours) & \textbf{0.201} & \textbf{85} & \textbf{17,226} \\
\hline
Compressed NAC & 0.207 & 5 & 3,445 \\
\hline
\end{tabular}
\caption{Comparison of mean Euclidean distance between predicted peak and to psuedo-voigt fitting on the test set.
Efficiency is measured in MegaBOPs (MBOPS).
Lower is better for all.
The best \textit{base} architecture in each category is bolded.
Our compressed architecture is quantized to 7 bits with 80\% weight sparsity.}
\label{tab:single100}
\end{table}


In our study, NAC compared to the current state-of-the-art in the field, BraggNN, as reported in \citet{BraggNN}.
To reiterate the context of our task, we train our model using images of Bragg peaks derived from a diffraction scan of an undeformed bi-crystal Gold sample, as detailed in \citet{dataset}.
This dataset comprises approximately 70,000 peaks across 1440 frames.
Peak locations from pseudo-Voigt fitting are used as the ground truth, which we optimize for in training via mean squared error.
For the evaluation of our model, the data is partitioned into a training, validation, and test set with an 80-10-10 split.  

To provide a fair comparison, each baseline architecture had their own hyperparameter optimization on the validation set with 100 trials for optimal training.
During evaluation, we measure performance by calculating the Euclidean distance between our approximation and the psuedo-Voigt label on the test set.
In addition, the distribution of the Euclidean distances on the test set is reported in Fig.~\ref{euclideandist}, showcasing that NAC matches the performance of BraggNN.

Once again, for efficiency BOPs is used as our main metric, with parameter count provided strictly for reference.
We note that while \citet{OpenHLS} has made some architecture changes, the main improvements of the paper is how they compile it on target hardware, a topic outside of the scope of this paper. We attempted to reimplement this architecture, but found poor convergence with mean distances $>4$.
We were able to find hyperparameters reporting $0.28$ distance on the validation set; however, these did not generalize, resulting in $5.5$ distance on the test set.
Therefore, this comparison was omitted from our results as we cannot reproduce the model.
 
\paragraph{Neural Architecture Analysis}
BraggNN employs a convolution-based attention unit, which exhibits quadratic complexity relative to the token length, set at 32 in their model.
This complexity creates a significant bottleneck in processing.
However, our global search revealed that this attention mechanism is not essential for achieving state-of-the-art performance.
Detailed architecture diagrams for each model variant are provided in the Appendix.
Instead of the conventional sequence of a conv-attention block followed by a standard convolutional block, our approach utilizes three consecutive convolutional blocks that progressively reduce the spatial dimensions.
Additionally, batch normalization is strategically incorporated earlier in the network, with layer normalization subsequently applied. It is important to note that while the global search provides a robust starting point, it is not the final step in optimization.
Our local search, focused on model compression, indicated that a significant portion of the weights in the larger convolutional layers are redundant for inference purposes.
This insight guided us in further refining the model, ensuring efficiency without compromising performance.

\section{Conclusion}
Our approach for neural architecture codesign addresses the challenges of using deep learning models effectively in science domains.
The global search phase in our pipeline was instrumental in identifying a model that not only excels in performance, but also improves efficiency.
Further optimization through local search, including additional HPO, pruning, and QAT significantly enhanced the model's efficiency.
This modular approach allows researchers and ML practitioners alike to optimize neural architectures without expert knowledge.
Since our metrics and search blocks are relatively plug and play, our method is domain agnostic.
We plan to prove this with future work, demonstrating the framework on a range of tasks such as jet classification and anomaly detection~\citep{Duarte:2022hdp}.

In future developments, the search space creation process will be streamlined by sampling a variety of pre-made model configurations to generate a useful block representation.
There is signficant room for improvement in our search strategy; we can fully exploit the hierarchical nature of the space with reinforcement learning.
The local search can also be optimized: mixed-precision quantization has not yet been explored, which is anticipated to enhance performance despite enlarging the search space.
We plan to deploy these models eventually on FPGAs, particularly for the CERN Large Hadron Collider Level-1 trigger~\cite{CMSL1T,ATLASL1T}.
The utilization of FPGA-specific tools like hls4ml~\citep{hls4ml} is expected to significantly boost the inference speed, as well.
This ongoing research promises to extend the utility of our approach to other scientific fields, enabling higher quality experimental results.

\section*{Acknowledgments}
NT and JD are supported by the U.S. Department of Energy (DOE), Office of Science, Office of Advanced Scientific Computing Research under the ``Real‐time Data Reduction Codesign at the Extreme Edge for Science'' Project (DE-FOA-0002501).
JD is also supported by the DOE, Office of Science, Office of High Energy Physics Early Career Research program under Grant No. DE-SC0021187, and the U.S. National Science Foundation (NSF) Harnessing the Data Revolution (HDR) Institute for Accelerating AI Algorithms for Data Driven Discovery (A3D3) under Cooperative Agreement No. OAC-2117997.
NT is also supported by the DOE Early Career Research program under Award No. DE-FOA-0002019.

\newpage
\appendix

\section{Appendix}
\subsection{Architectures}
Below, we display the architectures for the BraggNN model and the NAC base model.

\begin{lstlisting}[language=Python, caption={BraggNN model}]
Conv2d(1, 64, kernel_size(3x3), stride(1))
ConvAttn Block
    Q Weight: Conv2d(64, 32, 1x1, 1)
    K Weight: Conv2d(64, 32, 1x1, 1)
    V Weight: Conv2d(64, 32, 1x1, 1)
    Softmax()
    Projection: Conv2d(32, 64, 1x1, 1)
    LeakyReLU()
Conv Block
    Conv2d(64, 32, 3x3, 1)
    LeakyReLU()
    Conv2d(32, 8, 3x3, 1)
    LeakyReLU()
Flatten  
MLP
    Linear(200, 64)
    LeakyReLU()
    Linear(64, 32)
    LeakyReLU()
    Linear(32, 16)
    LeakyReLU()
    Linear(16, 8)
    LeakyReLU()
    Linear(8, 2)
\end{lstlisting}

\begin{lstlisting}[language=Python, caption={NAC base model}]
Conv2d(1, 32, kernel_size(3x3), stride(1))
Conv Block
    Conv2d(32, 4, 1x1, 1)
    ReLU()
    Conv2d(4, 32, 1x1, 1)
    BatchNorm2d(32)
    LeakyReLU()
ConvBlock
    Conv2d(32, 4, 1x1, 1)
    BatchNorm2d(4)
    GELU()
    Conv2d(4, 32, 3x3, 1)
    LayerNorm((32, 7, 7))
    GELU()
ConvBlock
    Conv2d(32, 8, 3x3, 1)
    LayerNorm((8, 5, 5))
    GELU()
    Conv2d(8, 64, 3x3, 1)
MLP
    Linear(576, 8)
    LayerNorm((8,))
    ReLU()
    Linear(8, 4)
    GELU()
    Linear(4, 4)
    LayerNorm((4,))
    GELU()
    Linear(4, 2)
\end{lstlisting}

\subsection{Search Space Details}

The hierarchical search space encompasses two distinct stages. The initial stage selects which pre-designed blocks will be incorporated into the model. This is followed by identifying the optimal hyperparameters specific to the chosen blocks.
For each block, the set of tunable parameters are detailed in Table~\ref{parameter_table}.
A common space for parameters, such as channel dimensions, is frequently utilized due to the repetitive nature of convolutional layers in this architecture.
As demonstrated in Table~\ref{parameter_table}, hyperparameters sampled for the Conv block offers about 200,000 combinations.
The attention block, restricted by the skip connection, has only 28 combinations due to the fixed kernel size and the need to maintain the input's spatial dimension.
Excluding the MLP, the search space for the three layers of blocks is roughly $10^{16}$.
Including the MLP classifier (also in Table~\ref{parameter_table} the total number of possible classifiers is approximately 2.5 million, bringing our entire search space to $10^{23}$.

\begin{table}[h]
\centering
\small
\renewcommand{\arraystretch}{1.2} 
\caption{Comprehensive parameter space values}
\begin{tabular}{|l|l|}
\hline
\textbf{Parameter} & \textbf{Space or description} \\
\hline
\multicolumn{2}{|c|}{\textbf{General parameter space}} \\
\hline
Block & \{Conv, Attention, None\} \\
Channel dimension & \{1, 2, 4, 8, 16, 32, 64\} \\
Kernel size & \{1, 3\} \\
Normalization method & \{Batch, Layer, None\} \\
Activation function & \{ReLU, GELU, LeakyReLU, None\} \\
Linear layer dimension & \{4, 8, 16, 32, 64\} \\
\hline
\multicolumn{2}{|c|}{\textbf{Conv block parameters}} \\
\hline
Conv1 in channels & Previous dimension \\
Conv1 out channels & Sample channel dimension \\
Conv1 kernel & Sample kernel size \\
Norm1 & Sample normalization method \\
Act1 & Sample activation function \\
Conv2 in channels & Sample channel dimension \\
Conv2 out channels & Sample channel dimension \\
Conv2 kernel & Sample kernel size \\
Norm2 & Sample normalization Method \\
Act2 & Sample activation function \\
\hline
\multicolumn{2}{|c|}{\textbf{Attention block parameters}} \\
\hline
QKV Dimension & Sample channel dimension \\
Skip Activation & Sample activation function \\
\hline
\multicolumn{2}{|c|}{\textbf{MLP classifier parameters} (all 4 layers)} \\
\hline
FC1 in dimension & Previous dimension \\
FC1 out dimension & Sample linear space \\
Norm1 & Sample normalization method \\
Act1 & Sample activation function \\
FC2 In Dimension & Previous dimension \\
FC2 Out Dimension & Sample linear space \\
Norm2 & Sample normalization method \\
Act2 & Sample activation function \\
\hline
\end{tabular}
\label{parameter_table}
\end{table}




\bibliography{aaai24}

\end{document}